\documentclass[Afour,sagev,times]{sagej}

\usepackage{moreverb,url}

\usepackage[colorlinks,bookmarksopen,bookmarksnumbered,citecolor=red,urlcolor=red]{hyperref}
\usepackage{subfigure}
\usepackage{mathtools}
\usepackage{multirow}
\usepackage{color}
\usepackage{xfrac}
\usepackage{ulem}

\newtheorem{remark}{Remark}
\newtheorem{definition}{Definition}
\newtheorem{lemma}{Lemma}
\newtheorem{theorem}{Theorem}

\newcommand\BibTeX{{\rmfamily B\kern-.05em \textsc{i\kern-.025em b}\kern-.08em
T\kern-.1667em\lower.7ex\hbox{E}\kern-.125emX}}

\def\volumeyear{2016}

\definecolor{LB}{RGB}{65,105,235}
\definecolor{LO}{RGB}{210,127,0}
\definecolor{LP}{cmyk}{0, 0.7808, 0.4429, 0.1412}
\definecolor{LG}{rgb}{0.13, 0.55, 0.13}
\definecolor{LR}{rgb}{0.77, 0.01, 0.2}
\definecolor{LV}{rgb}{0.5, 0.01, 0.5}

\begin{document}

\runninghead{Hegde, Aloor and Ghose}

\title{Bounded Distance-control for Multi-UAV Formation Safety and Preservation in Target-tracking Applications}

\author{Aditya Hegde\affilnum{1}\affilnum{*}, Jasmine Jerry Aloor\affilnum{2}\affilnum{*} and Debasish Ghose\affilnum{1}}

\affiliation{\affilnum{1}Indian Institute of Science, Bangalore, India\\
\affilnum{2}Indian Institute of Technology Kharagpur,
Kharagpur, India\\
\affilnum{*}Authors contributed equally to this work.}

\corrauth{Aditya Hegde, Department of Aerospace Engineering, Indian Institute of Science, Bangalore, 560012,
Karnataka, India.}

\email{adityahegde@iisc.ac.in}

\begin{abstract}
The notion of safety in multi-agent systems assumes great significance in many emerging collaborative multi-robot applications. In this paper, we present a multi-UAV collaborative target-tracking application by defining bounded inter-UAV distances in the formation in order to ensure safe operation. In doing so, we address the problem of prioritizing specific objectives over others in a multi-objective control framework. We propose a barrier Lyapunov function-based distributed control law to enforce the bounds on the distances and assess its Lyapunov stability using a kinematic model. The theoretical analysis is supported by numerical results, which account for measurement noise and moving targets. Straight-line and circular motion of the target are considered, and results for quadratic Lyapunov function-based control, often used in multi-agent multi-objective problems, are also presented. A comparison of the two control approaches elucidates the advantages of our proposed safe-control in bounding the inter-agent distances in a formation. A concluding evaluation using ROS simulations illustrates the practical applicability of the proposed control to a pair of multi-rotors visually estimating and maintaining their mutual separation within specified bounds, as they track a moving target.
\end{abstract}

\keywords{barrier Lyapunov function, {quadratic Lyapunov function, Robot Operating System, multi-objective control}, distributed control, formation control, target tracking, safety}

\maketitle
\section{Introduction}\label{intro}

\textbf{Unmanned Aerial Vehicles (UAVs)} are widely used in photogrammetry,\cite{tomasz2017} agriculture,\cite{christ2018,hegde2020}, search and rescue operations, \cite{dinnbier2017}, payload transportation,\cite{nasim2022,hegde2021cdc, lee2018} and target tracking applications. Recently, UAV-based target-tracking applications have received significant attention, and relevant work has been done to track moving objects visually.\cite{mohd2017} The object tracking problem is extended to cases with multiple moving objects, and the visual identification and tracking of UAVs have been performed by a camera mounted on another UAV.\cite{li2016} Collaborative object tracking, as discussed in Mueller et al. \cite{mueller2016} involves a team of UAVs tracking the object one at a time, depending on the battery charge. Unlike the application discussed in Mueller et al.,\cite{mueller2016} we consider a simultaneous object tracking problem involving a team of UAVs that hold their formation as they accomplish the target-tracking task. Such a formulation may be useful for a target localization task using the information from the multiple viewpoints that the UAVs provide. {Use of multiple UAVs also provides robustness to the object tracking and localization problem.}\cite{price2018,bhuvana2013,gohring2006} 

Applying multi-agent system theory to construct and operate UAV formations adds several benefits over operating individual UAVs by adding redundancy,\cite{schwalb2019} augmenting information availability,\cite{han2013} and reducing task completion time. The added complexity, however, (i) adversely affects system stability and performance characterization, and (ii) imposes additional constraints limiting complete utilization of the benefits of operating formations.

An imposition of agent interaction radius may lead to a time-varying interaction topology for the system,\cite{mor2005} which adds stochasticity to the prediction of system state evolution and stability analysis. Interaction and connectivity maintenance are thus essential for collaborative tasks, and control needs to be applied to the agents to preserve the connectivity of the network.\cite{sabat2015} In formations, these constraints are placed on inter-agent distances so that each agent can preserve its interaction with its neighbors.\cite{eg2001}

\begin{figure}[!t]
	\centering
	\subfigure[]{\includegraphics[width=0.49\linewidth]{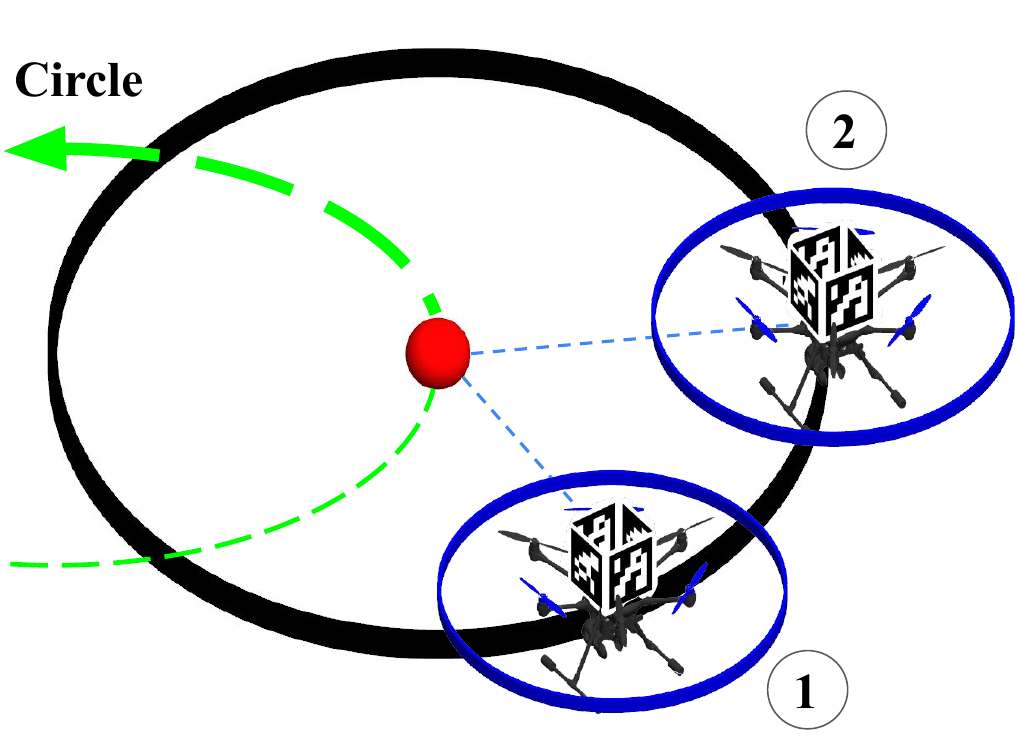}\label{fig:circle}}
	\subfigure[]{\includegraphics[width=0.49\linewidth]{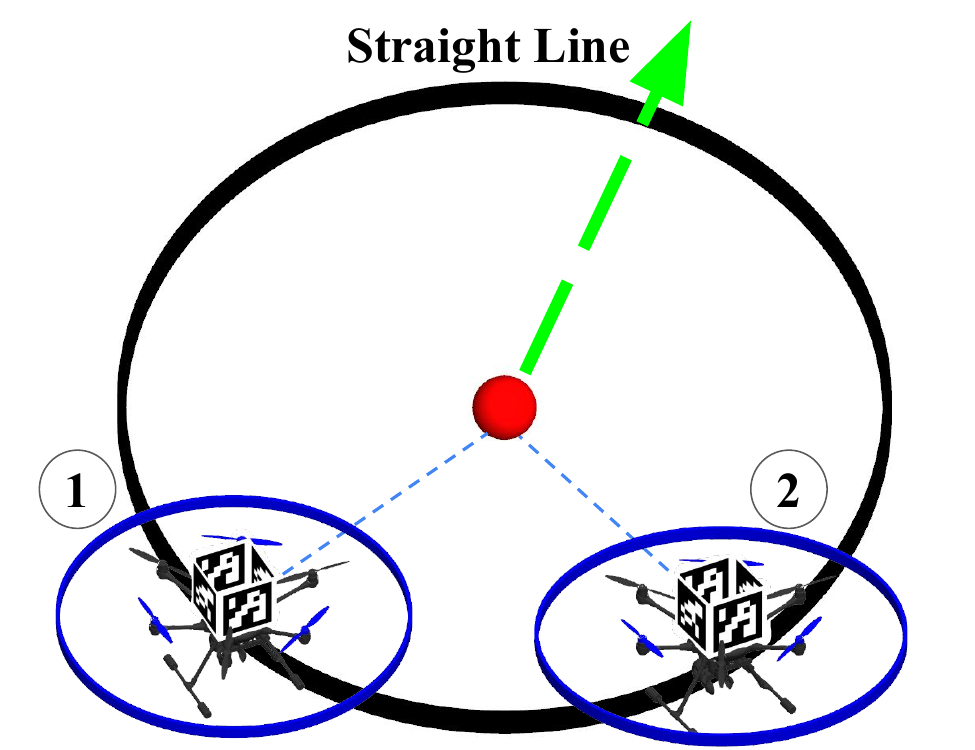}\label{fig:straight_line}}
	\caption{Summary of the approach: Two hexarotors track a continuously moving target while maintaining a distinct formation (their mutual separation). (a) Target following a circular trajectory, (b) Target following a straight-line trajectory.}
\end{figure}

Multi-objective multi-agent systems require the agent-collision avoidance objective to be included as a constraint in the formation control laws. One approach to enforcing the constraints in the formulation of control laws are Lyapunov-like barrier functions.\cite{panagou2016} Formation control and preservation in the presence of communication and measurement uncertainties are discussed in Han and Panagou.\cite{han2019} In line with the formation safety and preservation requirements of a collaborative target tracking application, we build on the idea of graph rigidity theory of formation control,\cite{zhao2019} and ensure safe operation by specifying bounds on inter-agent distances. We enforce these bounds using \textbf{barrier Lyapunov functions} \cite{tee2009barrier} (\textbf{BLFs}) and construct Lyapunov-like functions for a multi-objective problem using decentralized control. 

Collaborative target-tracking applications may require the UAVs to track targets farther away, and the Lyapunov function-based multi-objective formulations presented in this paper require one or a few objectives (formation safety and preservation) to be given a higher priority. Thus, existing approaches to Lyapunov function-based formation control \cite{zhang2016,keymasi2020} do not apply to problems with such requirements. Further, existing literature relies on assigning leader-follower roles to the agents in a formation, and creating virtual formations,\cite{oladapo2016} thus avoiding multi-objective control formulations for the agents. In contrast to this, we assume homogeneous roles for the agents in our work and present a multi-objective constrained control algorithm for each agent in the formation. The use of BLFs in our approach inherently yields a distributed distance preservation control law as compared to a centralized control formulation using \textbf{control barrier functions (CBFs)}.\cite{hegde2022} Further, it does not explicitly require an assumption to be made on the distance preservation efforts by the UAVs, as done in Alonso-Mora et al.\cite{mora2015} {The significant contributions of this paper are summarized below.}
\begin{enumerate}
	\item {The proposition of} a Lyapunov-like barrier function-based distributed-decentralized formation control law for applications that require minimal deviation of the inter-UAV distances from the reference distances, for safe operation of the formation.
	\item {The proposition of} a formation control law with a homogeneous treatment of the UAVs without any assumption on the contribution of each UAV towards the collective distance preservation objective.
	\item The inherent inclusion of constraints in the formation control law and the avoidance of solving a distributed optimization problem with coupled constraints.
\end{enumerate}
Results with a kinematic model of agents, as well as with hexa-rotor models in the \textbf{Robot Operating System (ROS)} environment, and accounting for control saturation are presented to demonstrate the practical applicability of the proposed formulation and associated control.

{The organization of the paper is as follows - the section \textit{System Description and Other Preliminaries} discusses the math preliminaries and the system model. We cover the objectives of the target tracking problem, propose the control, assess its stability, and provide numerical results with a kinematic model in the section \textit{Formation Safety and Preservation for Collaborative Target Tracking}. The section \textit{ROS Environment Simulation Results} presents ROS simulation results and practical applicability of the proposed approach. The paper is concluded with a discussion on the planned future work in the section \textit{Conclusions and Future Work}.} 

\section{System Description and Other Preliminaries}\label{prelim}

This section introduces the notations, system model, and definitions on barrier Lyapunov functions required for subsequent analysis. We use a multi-agent formation control approach and the UAVs are considered as agents in this section and the next (\textit{Formation Safety and Preservation for Collaborative Target Tracking}).

\subsection{Preliminaries}
The sets of real and non-negative real numbers are represented by $\mathbb{R}$ and $\mathbb{R}_+$, respectively. Vectors are represented by lowercase boldface and matrices by uppercase boldcase notations. A differentiable function $f : \mathcal{D} \to \mathbb{R}$, $\mathcal{D} \subseteq \mathbb{R}^n$ has a gradient $\nabla_{\pmb{x}} f = \left[\frac{\partial f}{\partial x_1}, \frac{\partial f}{\partial x_2}, \ldots, \frac{\partial f}{\partial x_n}\right]^T, \forall ~~\pmb{x}  = [x_1, x_2, \ldots, x_n]^T \in \mathcal{D}$, {where the $(\cdot)^T$ operator gives the transpose of the operand. The unit vector along the direction of a vector $\pmb{x}$ is represented as $\hat{\pmb{x}}$.} We use the $L_2$-norm in the subsequent sections and represent it by $\|\cdot\|$. We represent the inner product of $\pmb{x}_1,\pmb{x}_2 \in \mathbb{R}^n$ as $\pmb{x}_1 \cdot \pmb{x}_2$. The Cartesian product $\mathbb{R}^2\times\mathbb{R}^2,\ldots\times\mathbb{R}^2$ ($N$-times) is represented by $\mathbb{R}^{2N}$, where $\times$ represents the Cartesian product operator. We represent the formation of agents with an undirected graph $\mathcal{G} = (\mathcal{V},\mathcal{E})$, where the agents represent the elements of the finite set $\mathcal{V}$ and their interactions are represented as an unordered pair $(i,j) \in \mathcal{E} \subseteq \mathcal{V}\times\mathcal{V},~i,j~\in \mathcal{V}$. {A continuous function $f:[0,\infty) \rightarrow [0,\infty)$ is said to be a class $\mathcal{K}_\infty$ function if (i) it is strictly increasing, and (ii) $f(0) = 0$, (iii) $\lim_{r\rightarrow\infty} f(r) \rightarrow \infty$}.

\subsection{System Model}
We consider a system of $N$ agents moving in the $\mathbb{R}^2$ plane. The agents are holonomic, have first-order dynamics, and are represented by
\begin{equation}\label{sys_model}
\dot{\pmb{x}}_k = \pmb{v}_k = \pmb{u}_k , ~~~ k \in \{1,2,\ldots,N\} 
\end{equation}
{where, $\pmb{x}_k \in \mathbb{R}^2$,  $\pmb{v}_k \in \mathbb{R}^2$, and  $\pmb{u}_k \in \mathbb{R}^2$ are the position, velocity and control input for the $k^\text{th}$ agent in the inertial frame.} The control input $\pmb{u}_k$ in \eqref{sys_model} is further considered to be constrained as below

\begin{equation}\label{sat_control}
\begin{aligned}
\pmb{u}_k &= \text{sat}(\|\pmb{u}_k\|, u_{\text{max}})~\hat{\pmb{u}}_k\\
&=
\begin{cases}
\pmb{u}_k &  \text{if}~\|\pmb{u}_k\| \leq u_{\text{max}}\\
u_{\text{max}}~\hat{\pmb{u}}_k, & \text{if}~\|\pmb{u}_k\| > u_{\text{max}}
\end{cases}
\end{aligned}
\end{equation}
where, {$\hat{\pmb{u}}_k$ is the unit vector along the direction of $\pmb{u}_k$ and the $\text{sat}$ function saturates the magnitude $\|\pmb{u}_k\|$ to $u_{\text{max}} > 0$ along the direction of $\hat{\pmb{u}}_k$.}

This system of agents is a simple representation for a system of multi-rotors executing planar motion, with constraints placed on their velocity inputs, motivated by the PX4 autopilot \cite{PX4_meier} used for velocity-based control of multi-rotors (simulated in the ROS environment and discussed in the section \textit{ROS Environment Simulation Results}). The model is beneficial for preliminary stability analysis of the formation preservation algorithm presented here.
\begin{remark}
{The system model \eqref{sys_model} is chosen to align with the PX4 implementation of velocity reference-based \textbf{proportional-integral-derivative (PID)} control (used in the ROS simulations), and the BLF-based distance-bounding control proposed in this paper is extendable to systems modeled using second-order dynamics.}
\end{remark}

\subsection{Barrier Lyapunov Function}
Barrier Lyapunov functions (BLFs) \cite{tee2009barrier} provide the framework for ensuring safety through constrained system operation. They extend the idea of \textbf{control Lyapunov functions (CLFs)} by inherently including the constraints in the function and associated control law formulation. They have been extended to multi-objective, multi-agent problems requiring one or a few objectives to be performance-bounded.\cite{jain2019}

\begin{definition}[Barrier Lyapunov Function]\cite{tee2009barrier}
	A Barrier Lyapunov Function is a scalar function $V(\pmb{x})$ of state vector $\pmb{x} \in \mathcal{D}$ of the system $\dot{\pmb{x}} = f(\pmb{x})$ on an open region $\mathcal{D}$ containing the origin, that is continuous, positive definite, has continuous first-order partial derivatives at every point of $\mathcal{D}$, has the property $V(\pmb{x}) \rightarrow \infty$ as $\pmb{x}$ approaches the boundary of $\mathcal{D}$, and satisfies $V(\pmb{x}(t)) \leq \alpha, \forall t\geq 0$, along the solution of $\dot{\pmb{x}} = f(\pmb{x})$ for $\pmb{x}(0) \in \mathcal{D}$ and some positive constant $\alpha$.
\end{definition}

\begin{lemma}[Set invariance]\cite{tee2009barrier}\label{lem1}
	For any positive constants $k_{lo}$ and  $k_{hi}$, {a system with free states $\pmb{w} \in \mathbb{R}^l$ and a constrained state $z$}, let $\mathcal{Z} \coloneqq \{z \in \mathbb{R} \mid -k_{lo} < z < k_{hi}\} \subset \mathbb{R}$ and $\mathcal{N} \coloneqq \mathbb{R}^\ell  \times \mathcal{Z} \subset \mathbb{R}^{\ell+1}$ be open sets. Consider the {system dynamics} ${\dot{\pmb{\eta}} = \pmb{h}(t, \pmb{\eta})}$,  where $\pmb{\eta} \coloneqq [\pmb{w}^T, z]^T \in \mathcal{N}$, and ${\pmb{h} : \mathbb{R}_+ \times \mathcal{N}} \rightarrow \mathbb{R}^{\ell+1}$ is piecewise continuous in $t$ and locally Lipschitz in $z$, uniformly in $t$, on $\mathbb{R}_+ \times \mathcal{N}$. Suppose that there exist functions ${U: \mathbb{R}^\ell \rightarrow \mathbb{R}_+}$ and $V_1: \mathcal{Z} \rightarrow \mathbb{R}_+$, continuously differentiable and positive definite in their respective domains, such that $ V_1(z) \rightarrow \infty~~~\text{as}~~~ z \rightarrow -k_{lo}$ or $z \rightarrow k_{hi}$, and $\gamma_1(\|\pmb{w}\|) \leq U(\pmb{w}) \leq \gamma_2(\|\pmb{w}\|)$, where, $\gamma_1$ and $\gamma_2$ are class $\mathcal{K}_\infty$ functions. Let $V(\pmb{\eta}) \coloneqq V_1(z) + U(\pmb{w})$, and initial condition $z(0) \in \mathcal{Z}$. If $\dot{V} = (\nabla_{\pmb{\eta}} V)\cdot{\pmb{h}} \leq 0$ holds, then $z(t) \in \mathcal{Z},~\forall t \in [0, \infty)$, {and the set $\mathcal{Z}$ is forward invariant}.
\end{lemma}
\begin{proof}
    {The state space is split into the free states $\pmb{w}$ and the constrained state $z$. The barrier function $V_1(z)$ is constructed to restrict the state $z$ within the limits $-k_{lo}$ and $k_{hi}$ (and thus in the set $\mathcal{Z}$), while the function $U(\pmb{w})$ may be a quadratic Lyapunov function. Please refer to Lemma 1 and its proof in Tee et al. \cite{tee2009barrier} for details.}
\end{proof}

In the next section, we will use Lemma \ref{lem1} to construct a multi-objective Lyapunov-like function for each agent subjected to constraints.

\section{Formation Safety and Preservation for Collaborative Target Tracking}\label{prblm_sol}
The collaborative target tracking problem involves two aspects - target tracking by the individual agents, and formation safety and preservation. Formation safety is ensured by preventing inter-agent collisions, and preservation is necessary for maintaining communication between the agents for sharing and collecting target information. In this section, we identify these aspects as independent objectives and construct Lyapunov-like barrier functions. We then construct a multi-objective Lyapunov-like function and propose feedback control algorithms for the individual agents with the given information availability and control architecture.

\begin{figure*}[!t]
	\centering
	\subfigure[]{\includegraphics[width=0.6\columnwidth]{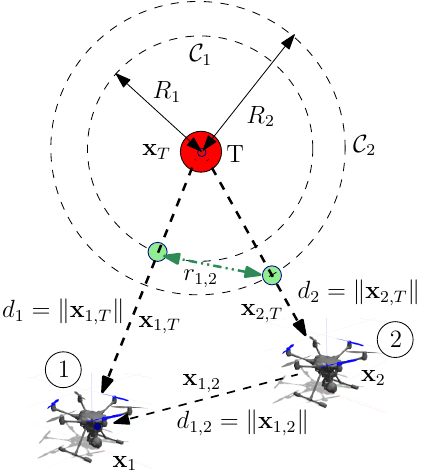}\label{Target_Tracking}}\hspace{1cm}
	\subfigure[]{\includegraphics[width=0.6\columnwidth]{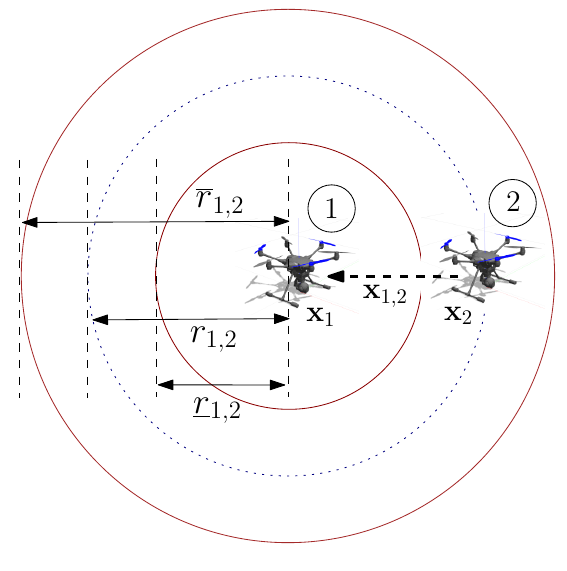}\label{Leader_Follower}}
	\caption{(a) Target tracking by two agents, with relevant inter-agent/target distances and desired configuration, (b) A two-agent formation with desired inter-agent distance ($r_{1,2}$), and associated lower ($\underline{r}_{1,2}$) and upper ($\overline{r}_{1,2}$) bounds.}
\end{figure*}

The target's position in the inertial frame is $\pmb{x}_T \in \mathbb{R}^2$. It is assumed that the target velocity, $\dot{\pmb{x}}_T = \pmb{v}_T$ is known. Each agent (for $k \in \{1,2,\ldots, N\}$) tracks the target, while maintaining its interaction with other agents (see Fig. \ref{Target_Tracking}).

\subsection{Target Tracking}
The distance of the $k^{\text{th}}$ agent from the target is $d_k = \|\pmb{x}_k - \pmb{x}_T\|=\|\pmb{x}_{k,T}\|$, and it is required to settle on a standoff circle $\mathcal{C}_k$ about the target, with a radius $R_k$. Fig. 2(a) shows a two-agent target-tracking problem with agents
settling on separate standoff circles. Such a case may arise when one of the agents may be required to be closer to the target than the other due to sensor limitations. {Let the tuple $\pmb{x} = (\pmb{x}_1, \pmb{x}_2, \ldots, \pmb{x}_N) \in \mathbb{R}^{2N}$ represent the states for the system of agents}. A Lyapunov function, $U(\pmb{x},\pmb{x}_T)$, is proposed for system \eqref{sys_model},
\begin{equation}\label{target_lyapunov}
U(\pmb{x},\pmb{x}_T)  = \sum_{k=1}^N U_k(\pmb{x}_k,\pmb{x}_T) = \frac{1}{4}\sum_{k=1}^N (d_k^2 -  R_k^2)^2,
\end{equation}
where, $U_k(\pmb{x}_k,\pmb{x}_T) = \frac{1}{4}(d_k^2 -  R_k^2)^2$ is the Lyapunov function for the $k^{\text{th}}$ agent. The time derivative of $U(\pmb{x},\pmb{x}_T)$ is
\begin{equation}\label{deriv_lyapunov}
\begin{aligned}
\dot{U}  &= \sum_{k=1}^N \dot{U}_k(\pmb{x}_k,\pmb{x}_T)\\ 
&= \sum_{k=1}^N \nabla_{\pmb{x}_{k,T}}U_k\cdot\dot{\pmb{x}}_{k,T}\\
&= \sum_{k=1}^N d_k(d_k^2 -  R_k^2)~\hat{\pmb{x}}_{k,T}\cdot\dot{\pmb{x}}_{k,T},
\end{aligned}
\end{equation}
where, $\hat{\pmb{x}}_{k,T}$ is the unit vector along the line joining the target to the agent, and the relation $\nabla_{\pmb{x}_{k,T}}U_k = d_k(d_k^2 -  R_k^2)\hat{\pmb{x}}_{k,T}$ is used to simplify the expression.

\subsection{Formation Safety and Preservation}
The formation safety and preservation problem requires the maintenance of inter-agent distances in the formation. However, the use of decentralized and distributed control in practical applications makes the maintenance of exact inter-agent distances difficult, requiring the placement of safety bounds (symmetric or asymmetric) on inter-agent distances. These bounds are included as constraints in the BLF associated with each pair of agents $(i,j) \in \mathcal{E}$. The BLF for the system associated with formation preservation is
\begin{equation}\label{blf_ia}
\begin{aligned}
V(\pmb{x}) &= \sum_{(i,j) \in \mathcal{E}}V_{i,j}(\pmb{x}_i,\pmb{x}_j)\\
&= \frac{1}{2}\sum_{(i,j) \in \mathcal{E}}\frac{(d_{i,j}^2 -  r_{i,j}^{2})^{2}}{(\overline{r}_{i,j}^{2} - d_{i,j}^2 )(d_{i,j}^2 -  \underline{r}_{i,j}^{2})},
\end{aligned}
\end{equation}
where {$V_{i,j}$ is the BLF for each unordered pair of agents $(i,j)$ in the set $\mathcal{E}$}, $d_{i,j} = \|\pmb{x}_i - \pmb{x}_j\| = \|\pmb{x}_{i,j}\|$ and $r_{i,j}$ is the desired inter-agent distance between each pair of interacting agents (see Fig. \ref{Target_Tracking}). The lower and upper bounds on the inter-agent distances for each pair are $\underline{r}_{i,j}$ and $\overline{r}_{i,j}$ (see Fig. \ref{Leader_Follower}). The time derivative of $V(\pmb{x})$ is
\begin{equation}\label{der_blf_ia}
\begin{aligned}
\dot{V} &= \sum_{(i,j) \in \mathcal{E}}\dot{V}_{i,j}(\pmb{x}_i,\pmb{x}_j)\\
&= \sum_{(i,j) \in \mathcal{E}} \nabla_{\pmb{x}_{i,j}}V_{i,j}\cdot\dot{\pmb{x}}_{i,j}\\
&= \sum_{(i,j) \in \mathcal{E}} \frac{ad_{i,j}\left(2b - a(\overline{r}_{i,j}^2 - 2d_{i,j}^2 + \underline{r}_{i,j}^2)\right)}{b^2}~\hat{\pmb{x}}_{i,j}\cdot\dot{\pmb{x}}_{i,j}
\end{aligned}
\end{equation}
where, $a = d_{i,j}^2 -  r_{i,j}^{2}$, $b = (\overline{r}_{i,j}^{2} - d_{i,j}^2 )(d_{i,j}^2 -  \underline{r}_{i,j}^{2})$ and $\nabla_{\pmb{x}_{i,j}}d_{i,j} = \hat{\pmb{x}}_{i,j}$ have been used to make the expression compact.
The preservation of the interaction topology and maintenance of a safe distance requires both agents in a pair $(i,j)$ to be active in maintaining the inter-agent distance within bounds.
We now propose a multi-objective control law using Lemma \ref{lem1}, and functions $U(\pmb{x},\pmb{x}_T)$ and $V(\pmb{x})$ for each agent in the formation.

\begin{table*}[tbhp]
\centering
\caption{Desired inter-agent distances and their bounds}\label{desdist_vals}%
\begin{tabular}{@{}lllll@{}}
\toprule
S.no. & Inter-agent distance & Desired value (m) & Lower bound (m) & Upper bound (m)\\
\midrule
1 & $r_{1,2}$ & $2$ & $1.8$ & $2.2$ \\
2 & $r_{2,3}$ & $2\sqrt{3} \approx 3.46$ & $3.2$ & $3.6$\\
3 & $r_{3,1}$ & $4$ & $3.8$ & $4.2$ \\
\bottomrule
\end{tabular}
\end{table*}

\begin{theorem}\label{theorem1}
	Let $\mathcal{G} = (\mathcal{V}, \mathcal{E})$ be the undirected and connected graph representing a formation of $N$ agents. Consider agent model \eqref{sys_model}, with initial states of the agents in the safe set $\mathcal{X}_r \coloneqq \{\pmb{x} \in \mathbb{R}^{2N} \mid \underline{r}_{i,j} < d_{i,j} < \overline{r}_{i,j}, ~\forall~(i,j) \in \mathcal{E}\}$, where $d_{i,j}$ is defined in \eqref{blf_ia}, and $\underline{r}_{i,j}, \overline{r}_{i,j} > 0$ are positive constants for $(i,j) \in \mathcal{E}$. Assuming that the target velocity $\pmb{v}_T$ is known, let the following saturated control law \eqref{sat_control} be applied to the agents 
	\begin{equation}\label{blf_control}
	\begin{aligned}
	\pmb{u}_k = &~\pmb{v}_T - K_T d_k\left(d_k^2 - R_k^2\right)\hat{\pmb{x}}_{k,T}\\
	&- K\sum_{j \mid (k,j) \in \mathcal{E}} \frac{a d_{k,j}}{b^2}\left\{2b - a(\overline{r}_{k,j}^2 - 2d_{k,j}^2 + \underline{r}_{k,j}^2)\right\}\hat{\pmb{x}}_{k,j}
	\end{aligned}
	\end{equation}
	where, $a$ and $b$ are as defined in \eqref{der_blf_ia}, and $\|\pmb{v}_T\| < u_{\text{max}}$ for all $k \in \{1,2,\ldots,N\}$. Then, the following statements hold:
	\begin{enumerate}
		\item[i)]If $K_T, K > 0$, all the agents asymptotically converge to the formation defined by $\underline{r}_{i,j} \leq r_{i,j} \leq \underline{r}_{i,j},~\forall~(i,j) \in \mathcal{E}$.
		\item[ii)] The trajectories of the agents stay within $\mathcal{X}_r$ for $t \geq 0$, that is, the safe set $\mathcal{X}_r$ is invariant for $t \geq 0$.	
	\end{enumerate}
\end{theorem}

\begin{proof}
	To prove the two statements together, we construct a combined Lyapunov-like function $W(\pmb{x}) = K_T U(\pmb{x},\pmb{x}_T) + K V(\pmb{x})$ for the system. The derivative of $W(\pmb{x})$ is
	\begin{equation}
	\begin{aligned}
	\dot{W} &= K_T\sum_{k=1}^N \nabla_{\pmb{x}_{k,T}}U_k\cdot\dot{\pmb{x}}_{k,T} + K\sum_{(i,j) \in \mathcal{E}}\nabla_{\pmb{x}_{i,j}}V_{i,j}\cdot\dot{\pmb{x}}_{i,j}\\
	&= \sum_{k=1}^N \bigg(K_T\nabla_{\pmb{x}_{k,T}}U_k + K\sum_{j \mid (k,j) \in \mathcal{E}}\nabla_{\pmb{x}_{k,j}}V_{k,j}\bigg)\cdot\dot{\pmb{x}}_{k,T},
	\end{aligned}
	\end{equation}
	where we use the relation $\dot{\pmb{x}}_{i,j} = \dot{\pmb{x}}_{i,T} - \dot{\pmb{x}}_{j,T}$. Using \eqref{blf_control} in $\dot{\pmb{x}}_{k,T} = \pmb{u}_k - \pmb{v}_T$ to simplify the expression, and substituting the expressions for $\nabla_{\pmb{x}_{k,T}}U_k$ and $\nabla_{\pmb{x}_{k,j}}V_{k,j}$ from \eqref{deriv_lyapunov} and \eqref{der_blf_ia}, we get $\dot{W} = - \sum_{k=1}^N \|\pmb{u}_k - \pmb{v}_T\|^2 \leq 0$ along the closed loop solutions of system \eqref{sys_model}, when $\|\pmb{v}_T\| < u_{\text{max}}$.
	
	The function $W(t) = K_T~U(t) + K~V(t)$ is finite and bounded by $W(t) \leq W(0) = K_T~U(0) + K~V(0), t \geq 0$ for finite initial distances of agents from the target $d_k(0),~\forall~k \in \{1,2,\ldots,N\}$ and initial positions of the agents in the safe set $\mathcal{X}_r$. Thus, the positions of agents remain in $\mathcal{X}_r$ for $t \geq 0$, proving (ii).
	
	To prove (i), we note that the formation defined by $\underline{r}_{i,j} \leq r_{i,j} \leq \underline{r}_{i,j},~\forall~(i,j) \in \mathcal{E}$ represents the set $\mathcal{X}_r$. Thus, the multiple equilibrium formations corresponding to $\pmb{u}_k = \pmb{v}_T$ and $\dot{W} = \dot{U} = \dot{V} = 0$ satisfy the specified distance bounds, for an initial formation that satisfies them.
\end{proof}

The proposed control for the $k^{\text{th}}$ agent thus requires distance and bearing information of the target ($T$) and its interacting agents $j$ such that $(k,j) \in \mathcal{E}$, along with information of the target velocity, $\pmb{v}_T$.

\subsection{{Comparison with Quadratic Lyapunov Functions}}
\textbf{Quadratic Lyapunov functions (QLF)} are widely used for multi-objective nonlinear control problems. Formation control approaches often utilize a QLF to satisfy the different formation requirements simultaneously. Each objective associated with the problem is encoded as a quadratic function of the system states, making the system Lyapunov function separable in its objectives. The existence of a QLF is sufficient to prove the stability of the proposed formation control laws. BLFs are a particular case of QLFs, and include system constraints in the structure of the separable quadratic functions. BLFs tend to infinity at the boundaries of a specified system operating set (see Lemma \ref{lem1}) instead of being radially unbounded like a QLF. In our comparison, we use a QLF of the form $W_Q(\pmb{x}) = K_T U(\pmb{x},\pmb{x}_T) + K Q(\pmb{x}),~ K_T,K > 0$, which is similar in its structure to the BLF we have proposed, comprising of the target tracking component $U(\pmb{x},\pmb{x}_T)$ (see \eqref{target_lyapunov}) and a formation preservation component $Q(\pmb{x})$. Here $Q(\pmb{x}) = \sum_{(i,j) \in \mathcal{E}}Q_{i,j}(\pmb{x}_i,\pmb{x}_j) = \frac{1}{4}\sum_{(i,j) \in \mathcal{E}}(d_{i,j}^2 -  r_{i,j}^{2})^{2}$ and the associated gradient-based QLF-control for the $k^{\text{th}}$ agent is 
\begin{equation}\label{qlf_control}
	\begin{aligned}
	\pmb{u}_k = &~\pmb{v}_T - K_T d_k\left(d_k^2 - R_k^2\right)\hat{\pmb{x}}_{k,T}\\
	&- K\sum_{j \mid (k,j) \in \mathcal{E}} d_{k,j}\left(d_{k,j}^2 - r_{k,j}^2\right) \hat{\pmb{x}}_{k,j}.
	\end{aligned}
\end{equation}
Such a QLF-control does not include constraints on the inter-agent distances and allows the formation to rearrange the cyclic order (topology) of its agents. This rearrangement leads to distance-equivalent formations \cite{zhao2019}. BLFs, on the other hand, ensure globally distance-rigid formations \cite{zhao2019} by allowing only formation translations and rotations. The comparison between QLFs and BLFs is summarized in Table \ref{diff_QLF}.

\begin{table}[tbhp]
\centering
\caption{Comparison bewteen QLFs and BLFs}\label{diff_QLF}%
\begin{tabular}{@{}p{0.5cm}p{3.5cm}p{3.5cm}@{}}
\toprule
S.no. & QLFs & BLFs\\
\midrule
\footnotesize 1 &  \footnotesize Positive semi-definite and equal to zero for the desired formation & \footnotesize Positive semi-definite and equal to zero for the desired formation\\
\footnotesize 2 &  \footnotesize Allow unconstrained motion of the agents & \footnotesize The inter-agent distance constraints are included in the function structure\\
\footnotesize 3 &  \footnotesize QLFs are radially unbounded & \footnotesize BLFs tends to infinity at the boundary of the system operating set\\
\footnotesize 4 & \footnotesize Cyclic order of agents in the formation is not unique due to the desired formation being defined as specified inter-agent distances & \footnotesize Though the desired formation is defined as in the case of QLFs, the added inter-agent distance constraints preserve the cyclic order of the agents in the formation\\
\footnotesize 5 & \footnotesize Leads to distance-equivalent\cite{zhao2019} formations in steady-state - unconstrained motion of agents is allowed & \footnotesize Leads to globally distance-rigid\cite{zhao2019} formations in steady-state - only translation and rotation of the formation, and limited motion of the agents about the initial/desired formation is allowed\\
\bottomrule
\end{tabular}
\end{table}

\subsection{{Numerical Simulation Results}}
We present simulation results with the BLF-control \eqref{blf_control} and QLF-control \eqref{qlf_control} applied to $N = 3$ agents using system model \eqref{sys_model} and the \textit{SciPy} library in \textit{Python}. The agents must settle on standoff circles $\mathcal{C}_1, \mathcal{C}_2$, and $\mathcal{C}_3$ about the target, with equal radii $R_1 = R_2 = R_3 = 2~\text{m}$. The desired inter-agent distances and the associated bounds are listed in Table \ref{desdist_vals}.

{The gains for both the controls are equally set to $K_T = 0.03$ and $K = 0.01$ for comparing their performance.} The bound on the control input to agents is $u_{\text{max}} = 3~\text{m/s}$. {We consider a straight-line and circular motion of the target and compare the results for the BLF and QLF-controls.} Further, we consider the effect of noise in the target velocity, agent-target, and inter-agent distance measurements used in the controls for the circular motion of the target.

\subsubsection{Straight-line motion of the target}
\begin{figure}[tbhp]
	\centering
	\vspace*{0.2cm}
	\subfigure[]{\includegraphics[width=0.49\linewidth]{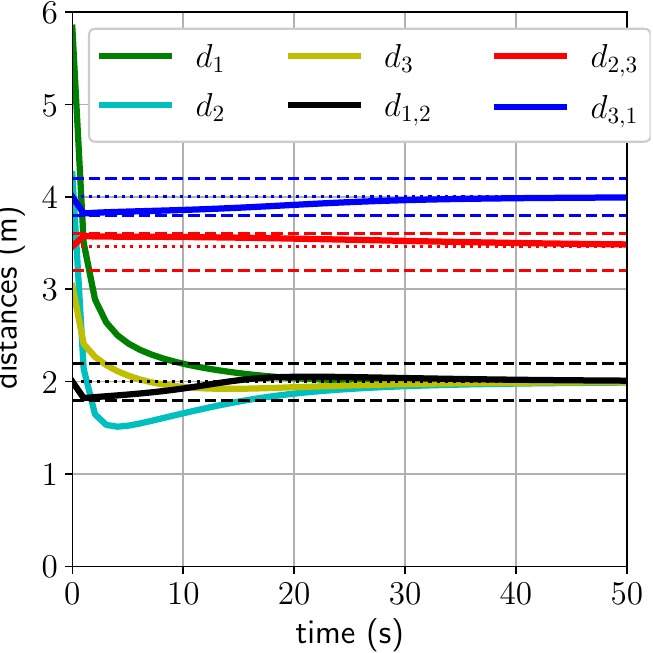}\label{blf_dist}}
	\subfigure[]{\includegraphics[width=0.49\linewidth]{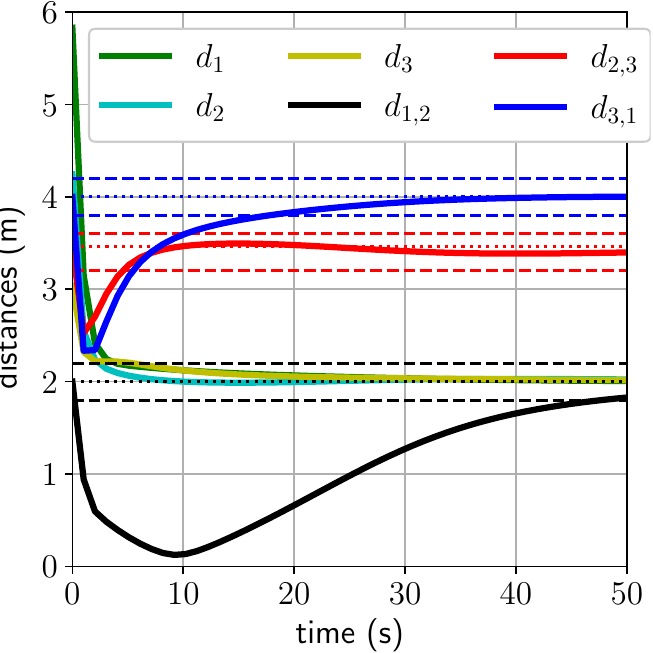}\label{qlf_dist}}
	\subfigure[]{\includegraphics[width=0.49\linewidth]{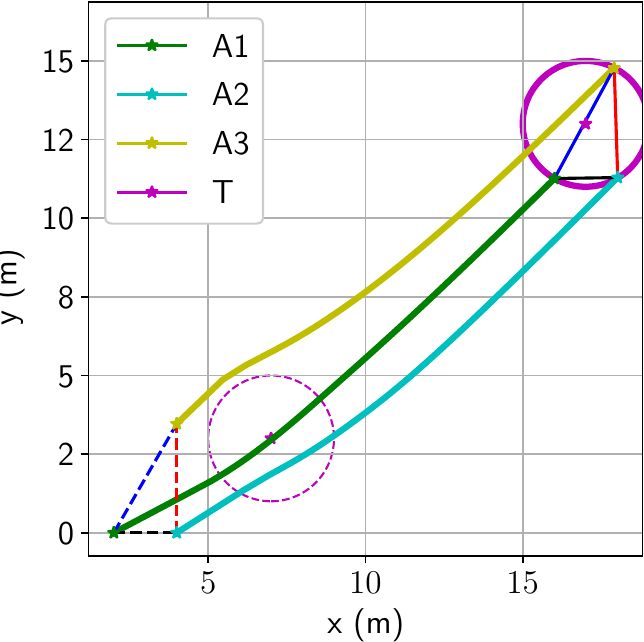}\label{blf_traj}}
	\subfigure[]{\includegraphics[width=0.49\linewidth]{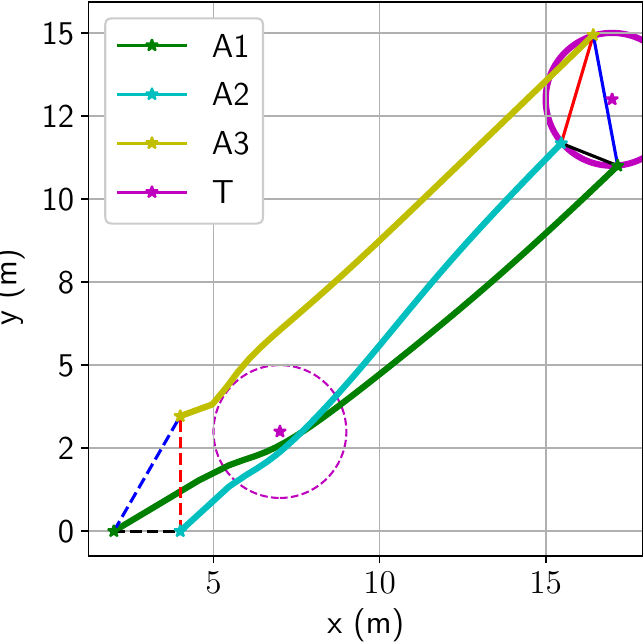}\label{qlf_traj}}
	\caption{A comparison of BLF and QLF-control for straight-line motion of target. Inter-agent and agent-target distances for (a) BLF-control and (b) QLF-control. Agent and target trajectories T - target; A1, A2, A3 - agents for (c) BLF-control (d) QLF-control}\vspace{-13pt}
\end{figure}

A moving target with velocity $\pmb{v}_T = [0.2, 0.2]^T \text{m/s}$ is considered. The initial inter-agent distances are at the desired values. The evolution of inter-agent distances and their associated bounds along with the agent-target distances are plotted in Fig. \ref{blf_dist}. A similar plot is presented for the QLF-control in Fig. \ref{qlf_dist}. It is observed that the inter-agent distances using BLF-control \eqref{blf_control} remain within the specified safe bounds (see $d_{1,2}, d_{2,3}, d_{3,1}$ in Fig. \ref{blf_dist}), while those in the case of QLF-control are unbounded (associated bounds are plotted in Fig. \ref{qlf_dist} for comparison with BLF-control). It is interesting to note the undershoot in the agent-target distances using BLF-control, which compensates for the bounded inter-agent distances, as compared to QLF-control.

\begin{figure}[tbhp]
	\centering
	\vspace*{0.2cm}
	\subfigure[]{\includegraphics[width=0.49\linewidth]{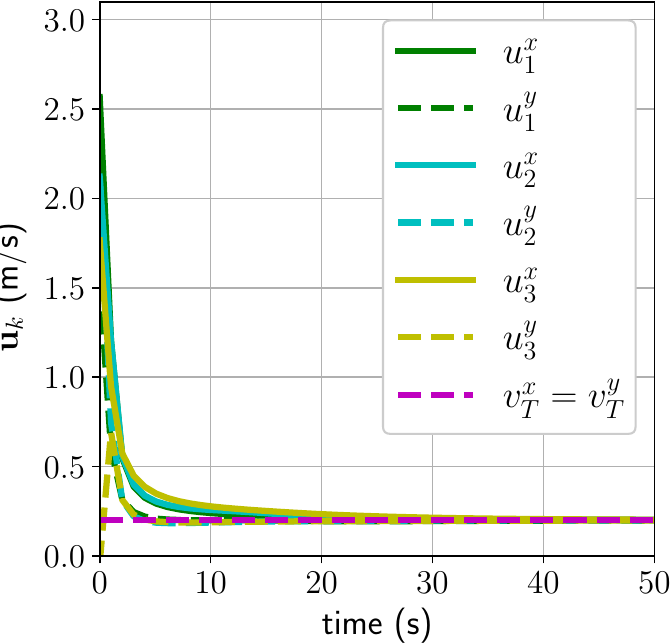}\label{blf_comp}}
	\subfigure[]{\includegraphics[width=0.49\linewidth]{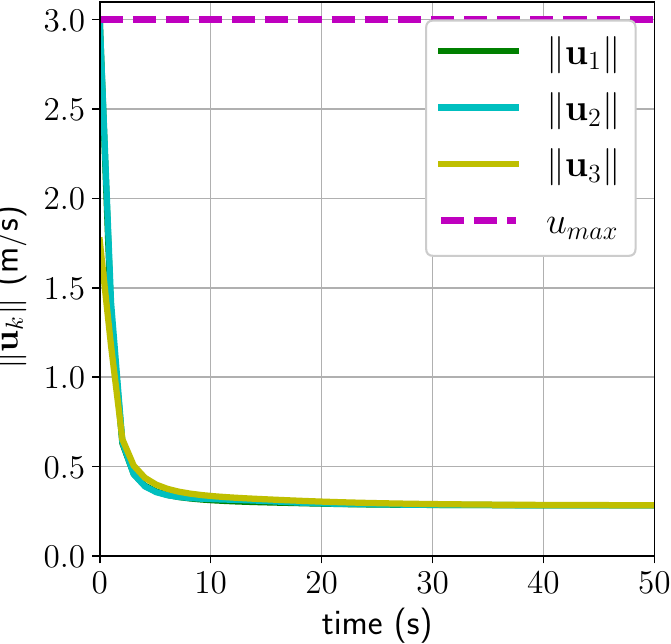}\label{blf_mag}}
	\caption{Control inputs for straight-line motion of target. (a) BLF-control components tracking target velocity $\pmb{v}_T = [0.2, 0.2]~\text{m/s}$ and (b) control magnitudes satisfying the bound $u_{\text{max}} = 3~\text{m/s}$}\label{ctrl_bound}\vspace{-13pt}
\end{figure}

The trajectories associated with the two control laws are plotted in Figs.\ref{blf_traj} and \ref{qlf_traj}. The initial and final positions of the triangular formation are plotted as dashed and solid lines. Similarly, the initial and final positions of the target and its associated standoff circle are plotted in dashed and solid magenta lines, respectively. The BLF-control ensures that the formation is pseudo-rigid and is capable of translating, rotating, and deforming but not rearranging (the cyclic order of the agents is unaffected). The agents may rearrange their cyclic order in the formation under QLF-control. Agent 2 moves across the line joining Agents 1 and 3 - the formation transitions through the condition $d_{1,2} + d_{2,3} = d_{3,1}$, seen in Fig. \ref{qlf_dist}. The distance between Agents 1 and 2 under QLF control decreases significantly (see Fig. \ref{qlf_dist}), which may lead to a collision if the physical dimensions of the agents are considered. The BLF control, on the other hand, ensures the agents' safety by bounding all the inter-agent distances.

The individual components of BLF-control and magnitudes for the agents are plotted in Fig. \ref{ctrl_bound}. The agent velocities (control inputs) asymptotically converge to the target velocity (see Fig. \ref{blf_comp}). The control inputs also satisfy the $u_{\text{max}}$ bound placed on them, as seen in Fig. \ref{blf_mag}. 

\subsubsection{Circular motion of the target}
The target moves on a circle of radius $R_T = 5$ m with a speed of $1$ m/s. We consider Gaussian noise with standard deviation of $0.02$ m/s in an agent's estimates of the target velocity components. The distance measurements are assumed to have noise with standard deviation of $0.02$ m.

\begin{figure}[t]
	\centering
	\vspace*{0.2cm}
	\subfigure[]{\includegraphics[width=0.49\linewidth]{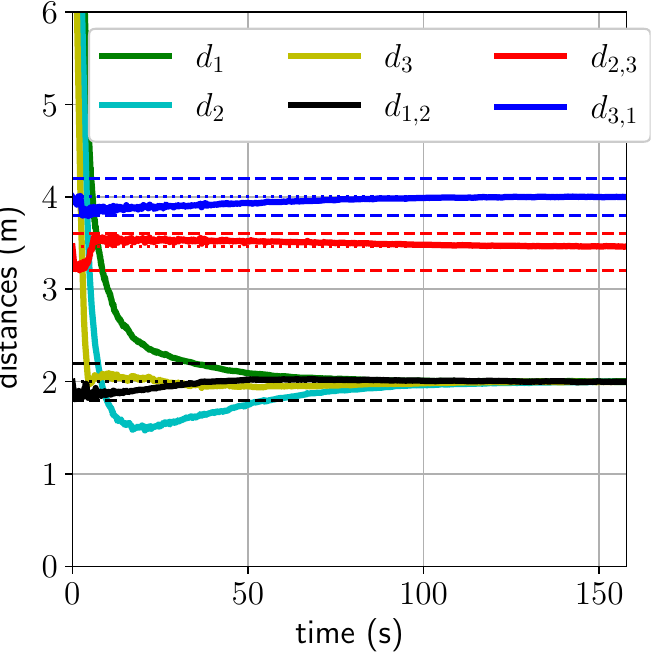}\label{blf_dist_circ}}
	\subfigure[]{\includegraphics[width=0.49\linewidth]{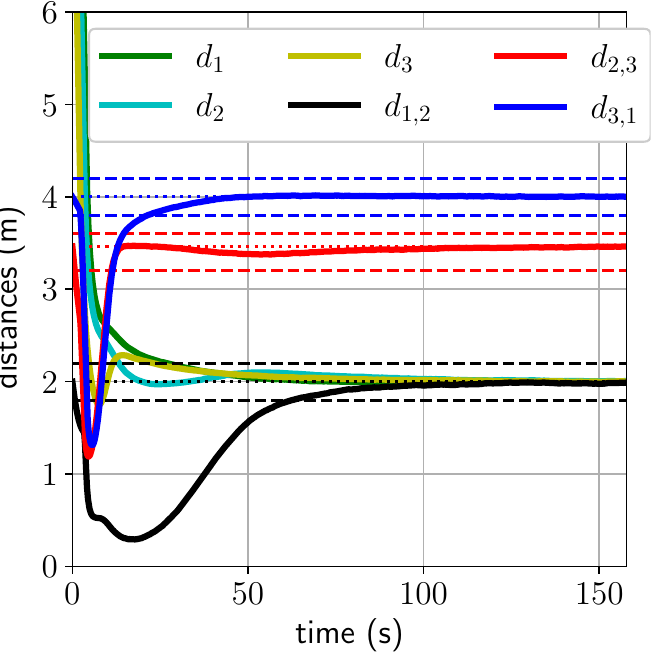}\label{qlf_dist_circ}}
	\subfigure[]{\includegraphics[width=0.49\linewidth]{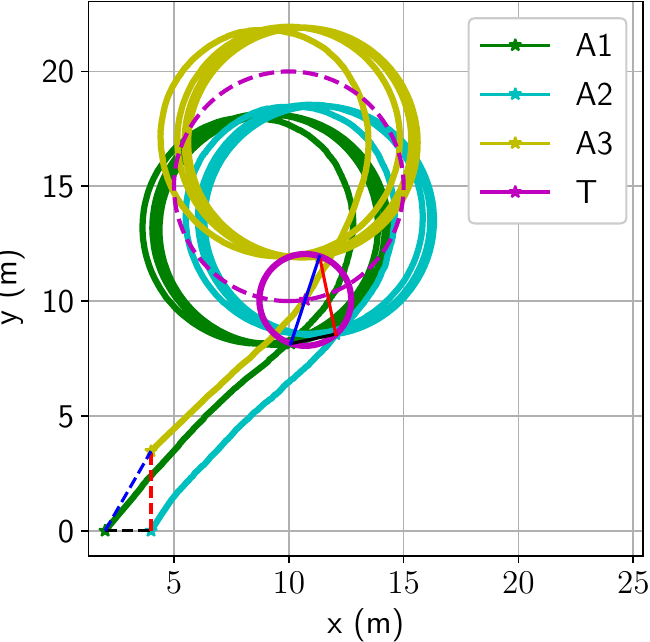}\label{blf_traj_circ}}
	\subfigure[]{\includegraphics[width=0.49\linewidth]{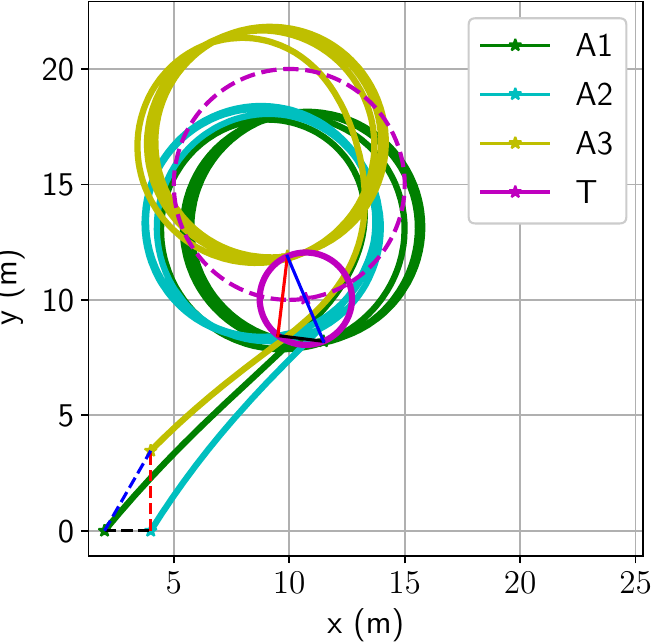}\label{qlf_traj_circ}}
	\caption{A comparison of BLF and QLF-control for circular motion of target. Inter-agent and agent-target distances for (a) BLF-control and (b) QLF-control. Agent and target trajectories T - target; A1, A2, A3 - agents for (c) BLF-control (d) QLF-control}\vspace{-13pt}
\end{figure}

\begin{figure}[t]
	\centering
	\vspace*{0.2cm}
	\subfigure[]{\includegraphics[width=0.49\linewidth]{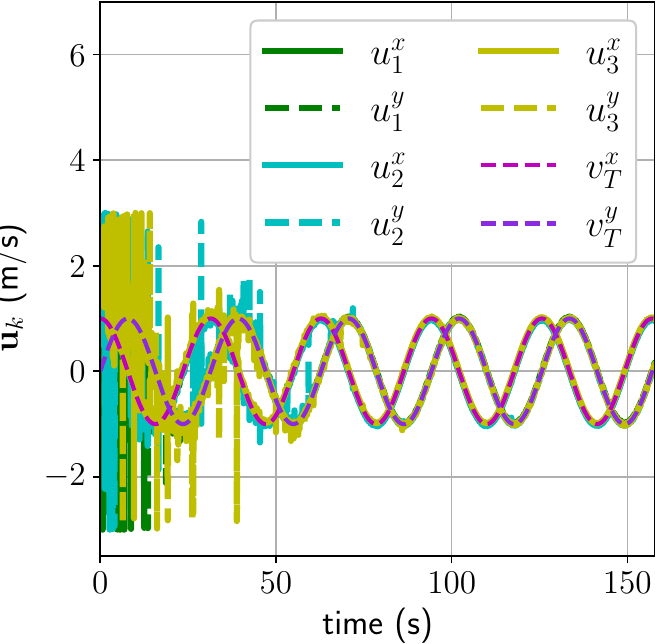}\label{blf_comp_circ}}
	\subfigure[]{\includegraphics[width=0.49\linewidth]{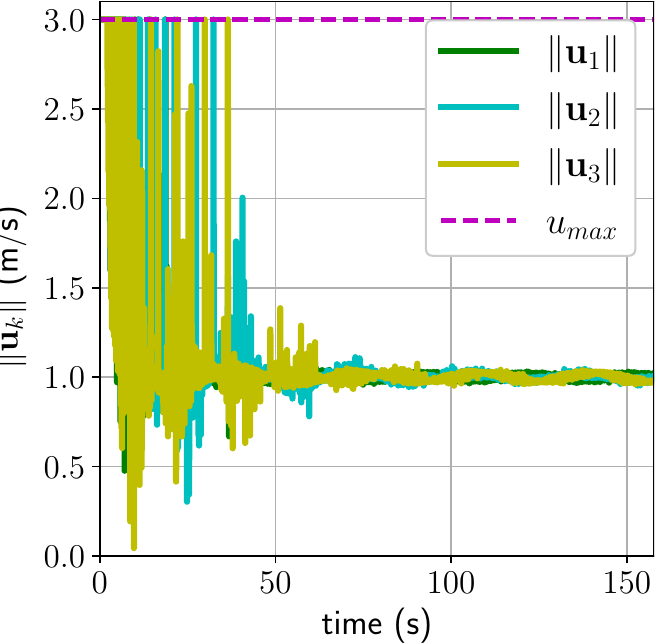}\label{blf_mag_circ}}
	\caption{Control inputs for circular motion of target. (a) BLF-control components tracking target velocity $\pmb{v}_T = [\cos 0.2t, \sin 0.2t]~\text{m/s}$ and (b) control magnitudes satisfying the bound $u_{\text{max}} = 3~\text{m/s}$}\label{ctrl_bound_c}\vspace{-13pt}
\end{figure}

As in the the case of straight-line motion of the target, the BLF-control ensures the agents' safety by restricting the deviation from the desired inter-agent distances within the specified bounds despite the noise in the target velocity and distance measurements (see Fig. \ref{blf_dist_circ}). Thus, the formation of agents does not rearrange its cyclic order, as seen in Fig. \ref{blf_traj_circ}, where the agents settle on the standoff circle about the target executing circular motion. The standoff circle and the circular trajectory of the target are marked by solid and dashed magenta lines, respectively. The three agents move on separate circles, with radii equal to $R_T$ but having different centers (Fig. \ref{blf_traj_circ}). The QLF-control also makes the three agents move on separate circles, but with changed cyclic order of the agents compared to the initial order (see Fig. \ref{qlf_traj_circ}). This reordering happens in the initial few seconds of the simulation, as is seen in Fig. \ref{qlf_dist_circ}, when Agents 1 and 2 are significantly close and the formation transitions through the condition $d_{1,2} + d_{2,3} = d_{3,1}$ (as in the case of straight-line motion of the target).

The control bound $u_{\text{max}}$ is satisfied, as seen in Fig. \ref{blf_mag_circ}. The control inputs to the agents display chatter due to noise in the measurements and operation of the system close to the specified distance bounds (see Figs.\ref{blf_comp_circ} and \ref{blf_mag_circ}). The chatter in the control can be correlated to the chatter in the inter-agent distances seen at the beginning of the simulation (in Fig. \ref{blf_dist_circ}). The control (and inter-agent distance) chatter eventually subsides as the inter-agent distances settle to their reference values and the agent velocities match the target velocity in Fig. \ref{blf_comp_circ}. The chatter can be avoided by filtering the distance measurements to implement the algorithm on the UAVs. Another critical aspect of designing the algorithm is suitably choosing the distance bounds, accounting for the noise in the velocity and distance measurements. A helpful metric for such an analysis is the least operating distance margin (difference of the desired inter-agent distances and their encompassing bounds),
\begin{equation}\label{dist-bnd-sel}
\gamma = \frac{\min_{(i,j)\in \mathcal{E}} \min (r_{i,j} - \underline{r}_{i,j}, \overline{r}_{i,j} - r_{i,j})}{SD_N},
\end{equation}
where $SD_N$ is the standard deviation of the noise in the inter-agent distance measurements. In our simulation $\gamma = \sfrac{0.14}{0.02} = 7$. A higher value of $\gamma$ is beneficial in increasing the robustness of the algorithm to noise.

\section{ROS Environment Simulation Results}\label{method_sec}
We now discuss the implementation of BLF-control to a pair of hexa-rotors in the ROS simulation environment and discuss its practical applicability.
\subsection{Simulation Environment}
The control \eqref{blf_control} is tested on the Robot Operating System using the Gazebo simulator with PX4 Software in the Loop (SITL) simulation. MAVROS is used to communicate with the models via ROS. The SITL mode provides greater flexibility for  testing various control algorithms in simulation before implementation on physical UAV models. A pair of Yuneec `Typhoon H480' hexa-rotors are used with each model mounted with four fiducial ArUco marker trackers on top of the hexa-rotors and visible from all four sides. Each model has a 2D lidar from the ROS rplidar package, and an RGB camera mounted on a gimbal. The lidar has a 360\textdegree\ field of view, a maximum range of 8 m, and a 5.5 Hz rotation frequency. The camera has 80\textdegree\ field of view and can be set to full 360\textdegree\ yaw using the gimbal. A custom Gazebo world is created to launch the two hexa-rotors and a target. The simulation is run at a rate of 60.0 Hz.


\subsection{Implementation}
The hexa-rotors are initialized at (0, 0, 0) m and (1, 0, 0) m, with an initial mutual distance of 1.0 m in the local coordinate system. We consider three cases: 
\begin{enumerate}
    \item Stationary target positioned at (5, 5, 4) m
    \item Target moving in a straight line with constant velocity
    \item Target moving in a circle with constant radius
\end{enumerate}

The hexa-rotors take off and hover at the specified altitude of $h=4~\text{m}$. The target's location and velocity are available to the control algorithm of the two hexa-rotors. The lidar data of each hexa-rotor provides the relative angular position of the other one. The camera of each hexa-rotor $k$ is then yawed to track the position of the other hexa-rotor, $j$, using the mounted marker. The marker’s translation and rotation vectors are determined using the OpenCV image processing library. Appropriate coordinate transformations \eqref{image_transformation} are performed to determine the position and distance of hexa-rotor $j$ relative to $k$ in the local frame of reference $\mathcal{U}$, using the camera on hexa-rotor $k$, $d^c_{k,j} = \|^\mathcal{U}\pmb{x}_{k,j}\|,~ k \in\{1,2\}$. 
\begin{equation}\label{image_transformation}
^{\mathcal{U}}\pmb{x}_{k,j}= ^\mathcal{U}_{b}\pmb{T} ~\ ^{b}_{f}\pmb{T} ~\ ^{f}_{C}\pmb{T} ~\  ^C\pmb{x}_{k,j}
\end{equation}
where, $^\mathcal{U}_{b}\pmb{T}$, $^{b}_{f}\pmb{T}$, and $^{f}_{C}\pmb{T}$ are the transformation matrices to transform distance from the hexa-rotor's body frame to local frame, fixed gimbal frame to the body frame and fixed gimbal frame to the camera's frame of reference, respectively. The position of Hexa-rotor $j$ in the frame of reference of the camera on $k$, is $^C\pmb{x}_{k,j}$, which is input to the algorithm. Each experiment has desired target standoff circles of $R_{k} \in (0.9,2.0)$ for $k \in \{1,2\}$, and the bounds on inter-hexa-rotor distance as $\overline{r}_{1,2} \in (1.0,4.0)$  and $\underline{r}_{1,2} \in (0.2,1.0)$ and $r_{1,2} = \frac{\underline{r}_{1,2} + \overline{r}_{1,2}}{2}$. The control generated by \eqref{blf_control} is input to the hexa-rotors as a reference velocity which the PX4 PID-velocity controller tracks \cite{PX4_meier}.

\begin{figure}[h!]
	\centering
	\subfigure[]{\includegraphics[width=0.49\linewidth]{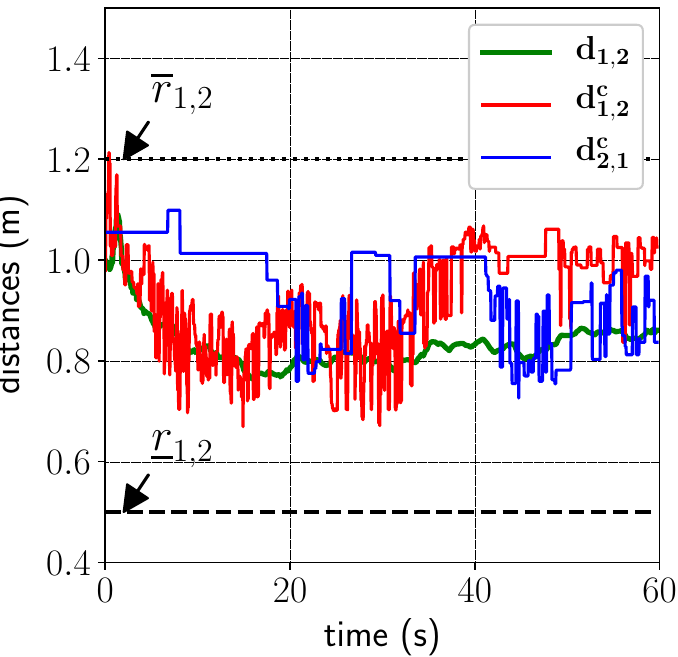}\label{inter_ag_dist}}
	\subfigure[]{\includegraphics[width=0.49\linewidth]{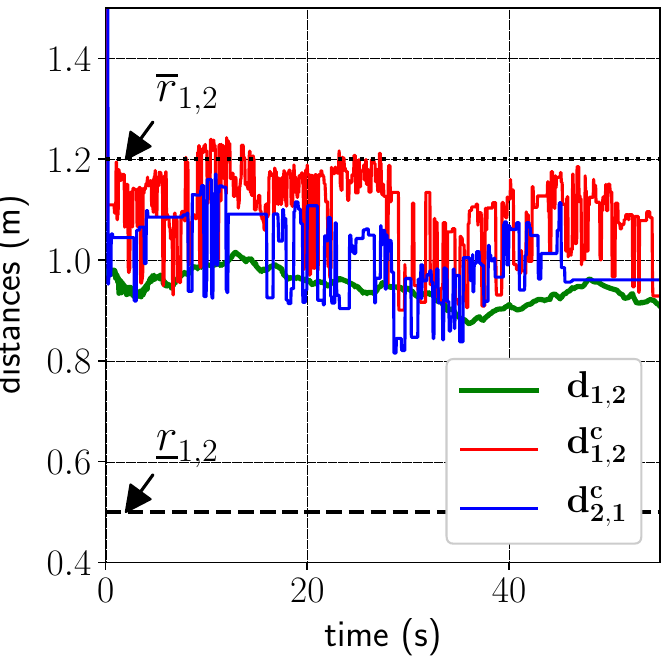}\label{qlf_inter_ag_dist}}
	\subfigure[]{\includegraphics[width=0.49\linewidth]{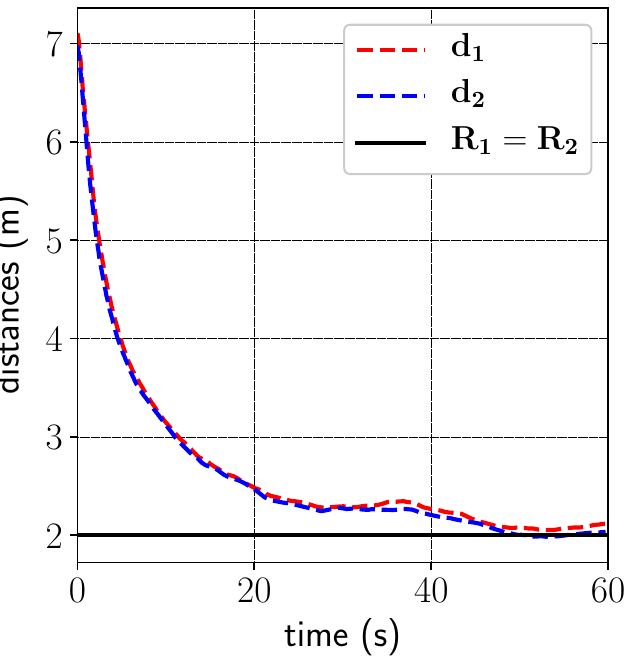}\label{target_ag_dist}}
	\subfigure[]{\includegraphics[width=0.49\linewidth]{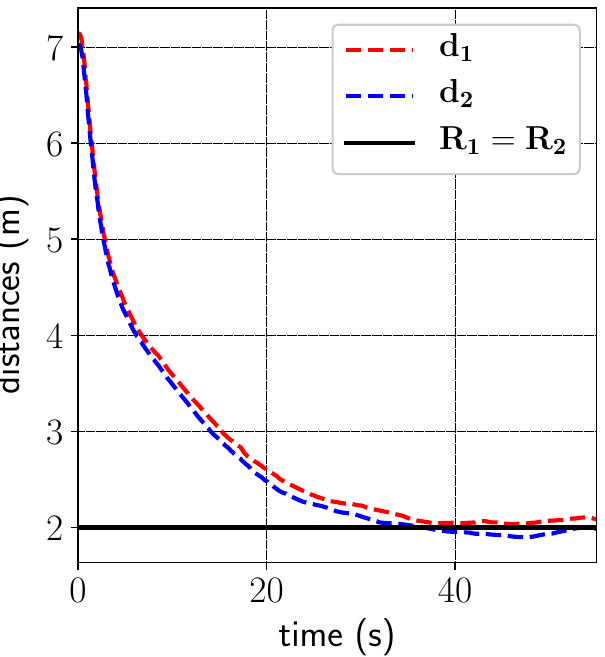}\label{qlf_target_ag_dist}}
	\subfigure[]{\includegraphics[width=0.49\linewidth]{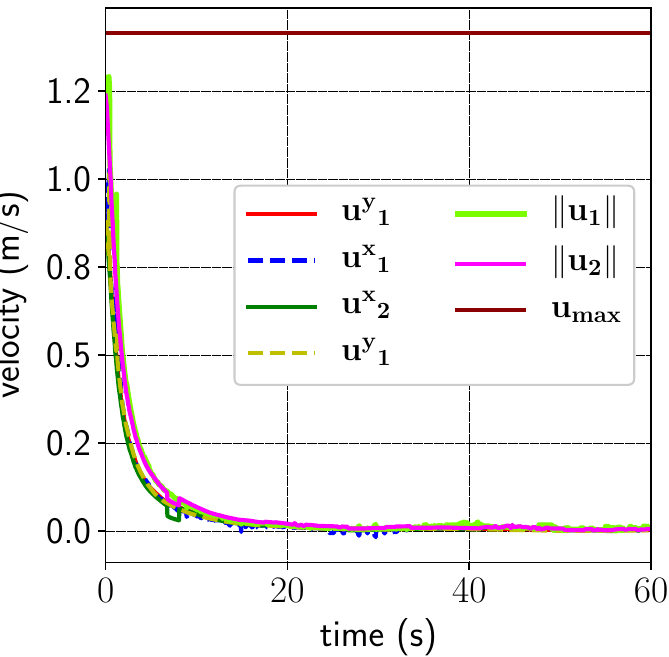}\label{inter_ag_vel}}
	\subfigure[]{\includegraphics[width=0.49\linewidth]{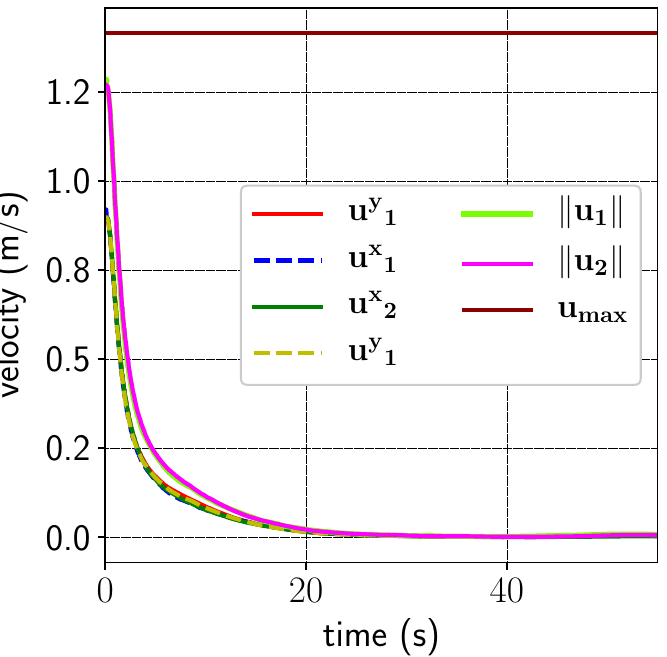}\label{qlf_inter_ag_vel}}
	\caption{A comparison of BLF and QLF-control for a stationary target. Actual and estimated inter-hexa-rotor distances for (a) BLF-control, (b) QLF-control. Hexa-rotor-target distances for (c) BLF-control, (d) QLF-control. Control inputs (velocities) components and magnitude for (e) BLF-control, (f) QLF-control}\label{ROS_sim}
\end{figure}

\begin{table*}[h]
\begin{center}
\begin{minipage}{\textwidth}
\caption{Summary of results for inter-hexa-rotor bounds and gain variation, $R_1=R_2 = 2~\text{m}$ for a stationary target}\label{tab1}
\begin{tabular*}{\textwidth}{@{\extracolsep{\fill}}cccccp{7cm}}
\toprule%
& \multicolumn{1}{@{}c@{}}{Hex-Target} & \multicolumn{3}{@{}c@{}}{Inter-Hex}& \\\cmidrule{2-2}\cmidrule{3-5}%
Sl.no. &  $K_T$ & \textbf{$\underline{r}_{i,j}$} & \textbf{$\overline{r}_{i,j}$}& $K$ & Observations \\
\midrule
			1 & 0.02 & 0.5 & 2.0 & 0.004 & \multirow{2}{=}{Inter-hexa-rotor distance stays within bounds, fast convergence to desired distances}\\
			2 & 0.02 & 0.5 & 1.3 & 0.004 & \\
			3 & 0.02 & 0.5 & 1.2 & 0.004 & System does not stay within bounds, does not converge to the the desired distances\\
			4 & 0.004 & 0.5 & 1.2 & 0.002 & Inter-hexa-rotor distance stays within bounds, converges to the desired distances slowly due to the reduced gains\\
\bottomrule
\end{tabular*}
\end{minipage}
\end{center}
\end{table*}

\begin{table*}[h]
\begin{center}
\begin{minipage}{\textwidth}
\caption{Summary of results for inter-hexa-rotor bounds and gain variation,  $R_1=R_2 = 3~\text{m}$,  for a target moving in a straight line}\label{tab2}
\begin{tabular*}{\textwidth}{@{\extracolsep{\fill}}ccccccp{4.8cm}}
\toprule%
& \multirow{1}{*}{$\pmb{v}_T$} &  \multicolumn{3}{c}{{Inter-Hex}} & {Hex-Target} & \\\cmidrule{2-2}\cmidrule{3-5}\cmidrule{6-6}%
Case & m/s & \textbf{$\underline{r}_{i,j}$} & \textbf{$\overline{r}_{i,j}$} & $K$ & $K_T$ & Observations \\
\midrule
			1 & 0.11 & 0.5 & 2.0 & 0.0075 &   0.015 & 
			{Inter-hexa-rotor distance stays within bounds, fast convergence to desired distances}\\
            2 & 0.11 & 0.5 & 1.3 & 0.005 &   0.001 & 
			 {Inter-hexa-rotor distance stays within bounds, very slow convergence to desired distances}\\
			
			3 & 0.17 & 0.5 & 2.0 & 0.045 &   0.0175 & {System exhibits oscillatory behaviour about the desired target-hex position, while maintaining inter-hex distances}\\
		
			4 & 0.17 & 0.5 & 1.3 & 0.001 &   0.0012 & {System exhibits oscillatory behaviour about the desired target-hex position, very slow convergence to desired distances}\\
\bottomrule
\end{tabular*}
\end{minipage}
\end{center}
\end{table*}

\begin{table*}[h]
\begin{center}
\begin{minipage}{\textwidth}
\caption{Summary of results for inter-hexa-rotor bounds and gain variation,  $R_1=R_2 = 3~\text{m}$,  for a circularly moving target}\label{tab3}
\begin{tabular*}{\textwidth}{@{\extracolsep{\fill}}ccccccp{4.8cm}}
\toprule%
& \multirow{1}{*}{$\pmb{v}_T$} &  \multicolumn{3}{c}{{Inter-Hex}} & {Hex-Target} & \\\cmidrule{2-2}\cmidrule{3-5}\cmidrule{6-6}%
Case & m/s & \textbf{$\underline{r}_{i,j}$} & \textbf{$\overline{r}_{i,j}$} & $K$ & $K_T$ & Observations \\
\midrule
			1 & 0.17 & 0.5 & 2.0 & 0.015 & 0.015 & 
			 {Inter-hexa-rotor distance stays within bounds, fast convergence to desired distances}\\
			
			2 & 0.17 & 0.5 & 1.3 & 0.0036 &   0.015 & {Inter-hexa-rotor distance stays within bounds, slow convergence to desired distances}\\
		
			3 & 0.6 & 0.5 & 2.0 & 0.045 &   0.0125 & {System exhibits oscillatory behaviour about the desired target-hex position, while maintaining inter-hex distances}\\

			4 & 0.6 & 0.5 & 1.3 & 0.0036 & 0.0125 & {System exhibits oscillatory behaviour about the desired target-hex position, slow convergence to inter-hex distances}\\
\bottomrule
\end{tabular*}
\end{minipage}
\end{center}
\end{table*}

\subsection{Experiments and Results}
We present a few results for the first case where the target is kept stationary. With the initial conditions specified above, we vary the inter-hexa-rotor safe distance bounds starting with ($\underline{r}_{1,2}, \overline{r}_{1,2}$) = (0.2, 2.0) and gains $K$ = 0.004, $K_T$ = 0.02. The outer bound is decreased in steps of 0.1 m till 1.3 m. For further decrease in the higher bound, $K$ is decreased to 0.002 to ensure that the hexa-rotors stay within the bounds. For any further increase in the lower bound, it is required to reduce $K_T$ to 0.004. A few results of varying the bounds are summarised in Table \ref{tab1}. We observe that tighter bounds (see Case 3, Table \ref{tab1}) lead to higher control inputs to the hexa-rotors due to the barrier function, eventually leading to the transgression of these safe bounds and oscillatory behavior. It is thus required to tune the gains $K_T$ and $K$ accordingly, which may lead to a slower response (see Case 4, Table \ref{tab1}). The results from Case 4 are presented in Fig. \ref{ROS_sim} along with results of applying QLF-control to the same problem. It is observed that the BLF-control maintains the inter-hexa-rotor distance within bounds and also leads to a faster convergence to the desired inter-hexa-rotor distance of $r_{1,2} = 0.85~\text{m}$, as compared to QLF-control (see Figs.\ref{inter_ag_dist} and \ref{qlf_inter_ag_dist}) for the same gain values. {For finite bounds and similar gains for QLF and BLF-based control, the BLF-control exhibits higher control efforts and faster response than the QLF-control. This behavior is attributed to the effect of bounds on the control magnitude, which is inversely proportional to the distances from the associated bounds.} The low gain values required for keeping the inter-hexa-rotor distance bounded adversely affect the hexa-rotor-target distance tracking. However, this problem may be addressed with gain-scheduling.

\begin{figure}[!h]
	\centering
	\subfigure[]{\includegraphics[width=0.49\linewidth]{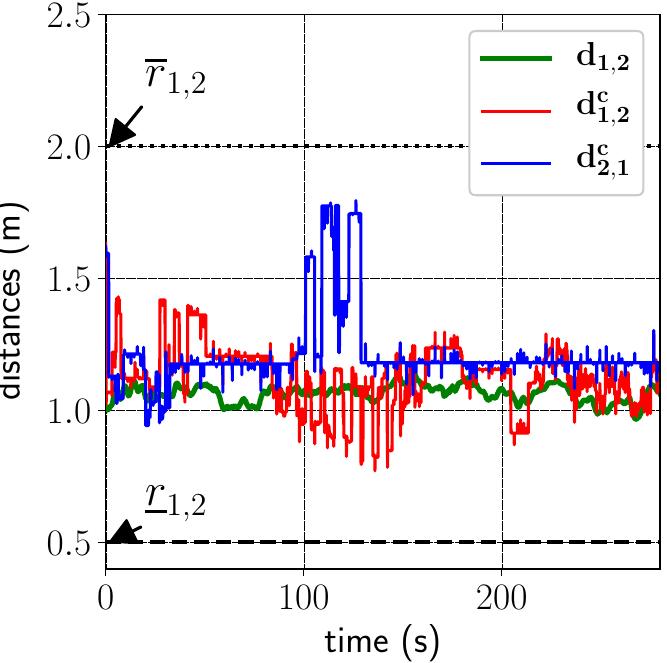}\label{inter_ag_dist_c}}
	\subfigure[]{\includegraphics[width=0.49\linewidth]{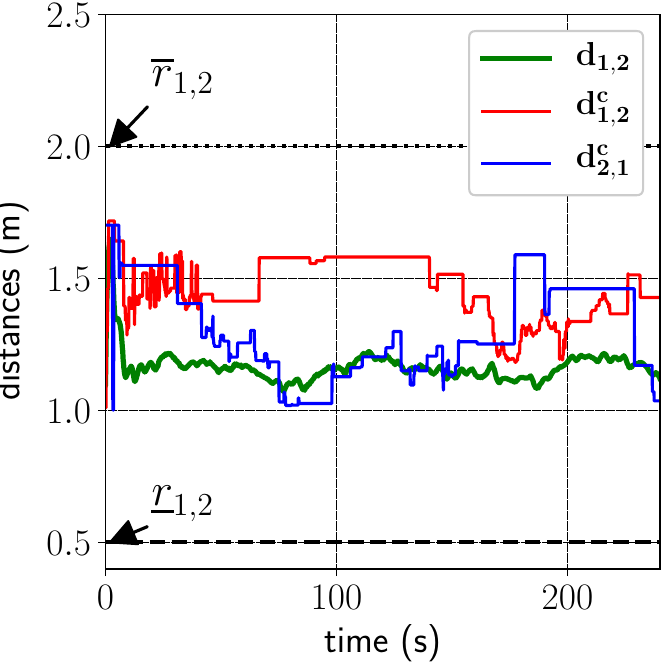}\label{inter_ag_dist_s}}
	\subfigure[]{\includegraphics[width=0.49\linewidth]{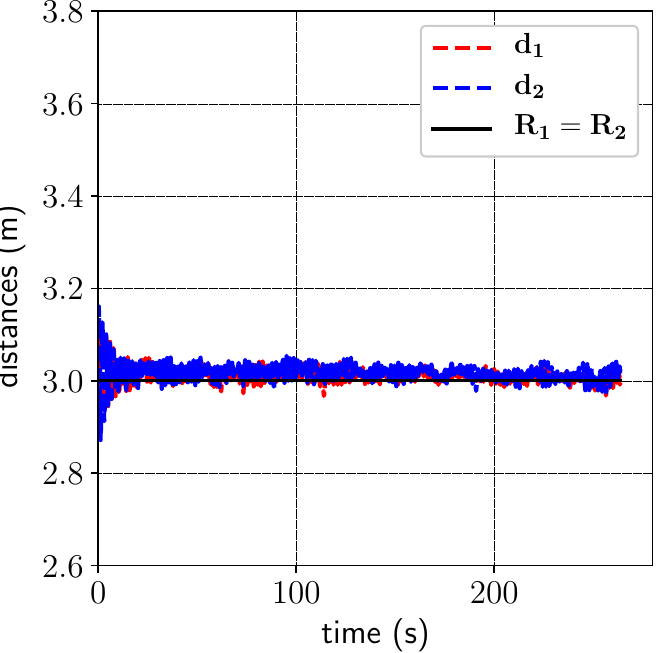}\label{target_ag_dist_c}}
	\subfigure[]{\includegraphics[width=0.49\linewidth]{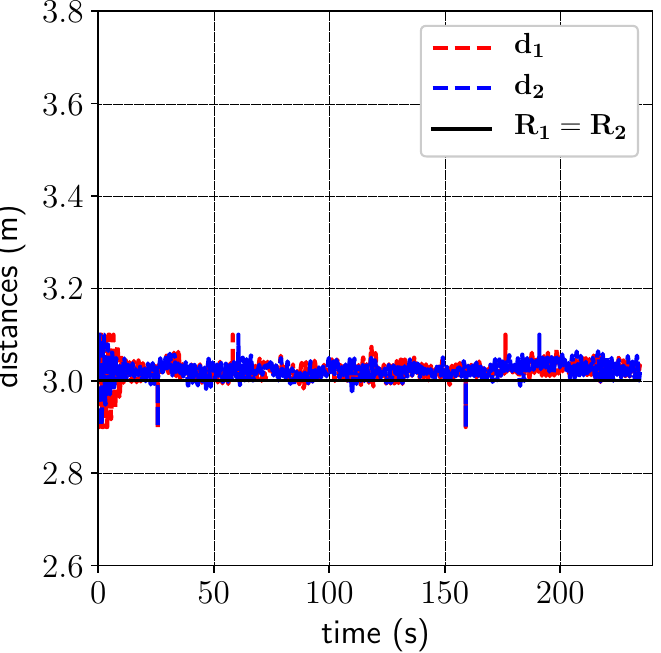}\label{target_ag_dist_s}}
	\subfigure[]{\includegraphics[width=0.49\linewidth]{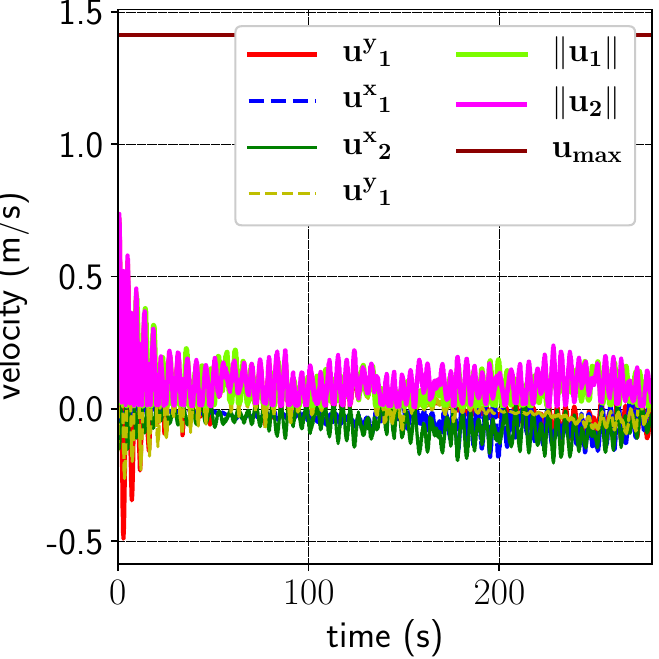}\label{inter_ag_vel_c}}
	\subfigure[]{\includegraphics[width=0.49\linewidth]{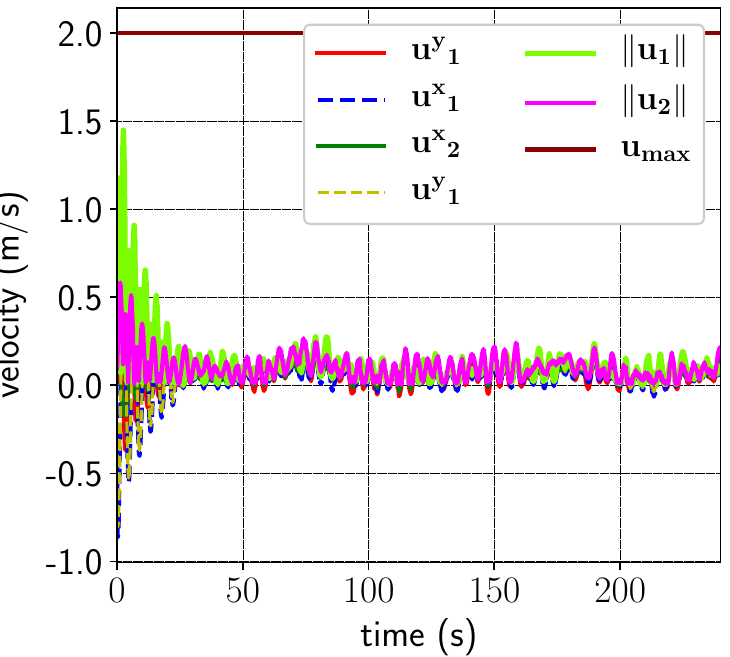}\label{inter_ag_vel_s}}
\caption{BLF-control for circular and straight-line motion of the target. Actual and estimated inter-hexa-rotor distance for (a) circular motion , (b) straight-line motion. Hexa-rotor-target distance for (c) circular motion, (d) straight-line motion. Control inputs (velocities) components and magnitudes for (e) circular motion, (f) straight-line motion}\label{ROS_sim_circle_straight}
\end{figure}

\begin{figure}[!h]
	\centering
    \subfigure[]{\includegraphics[width=0.48\linewidth]{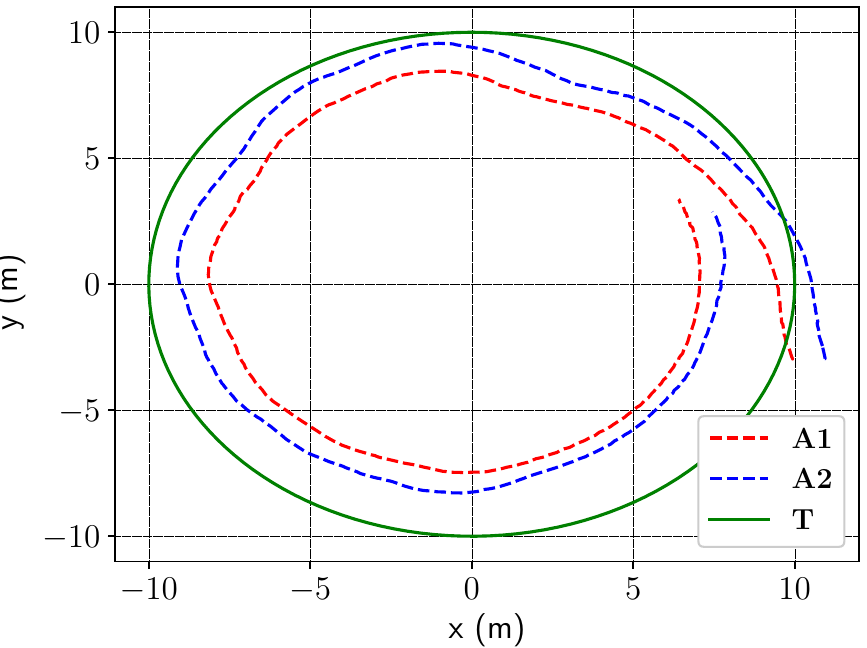}\label{c_traj}}
	\subfigure[]{\includegraphics[width=0.48\linewidth]{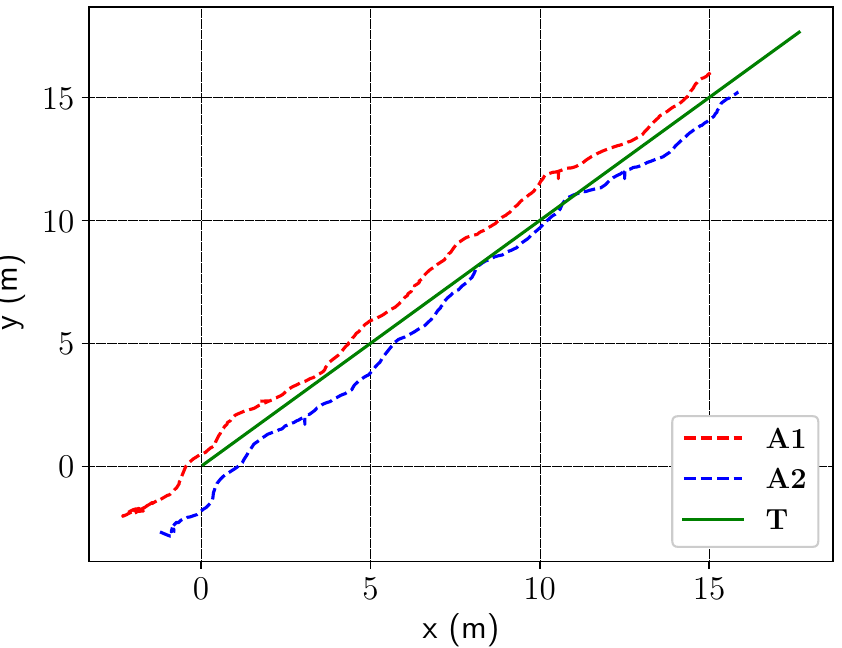}\label{s_traj}}
	\caption{Hexa-rotor (A1, A2) and target (T) trajectories for (a) circular motion, and (b) straight-line motion}\label{str-circ-ros-traj}
\end{figure}

We extend the tests for a stationary target to a target moving with constant velocity and a target moving along a circular trajectory. We vary the inter-hexa-rotor distance safe bounds starting with ($\underline{r}_{1,2}, \overline{r}_{1,2}$) = $(0.5, 2.0)$ m and gains $K = 0.0075, K_T = 0.015$. The target has a speed of $0.11$ m/s along the straight line $y=x$. For circular motion, the target moves with a speed of $0.17$ m/s and tracks a circle of radius $10$ m. The hexa-rotors are required to maintain a distance of $3.0$ m from the target. We summarise the results for the case of straight-line and circular motion of the target in Tables \ref{tab2} and \ref{tab3}, respectively. We observe that the hexa-rotors strictly maintain the desired hexa-rotor-target distances along with the inter-hexa-rotor distance when the target is moving at this velocity (see Case 1, Tables \ref{tab2}, \ref{tab3}). When we decrease the outer bound to $1.3$ m, we reduce the gains $K$ and $K_T$ to lower values to ensure that the hexa-rotors maintain desired distances (see Case 2, Tables \ref{tab2}, \ref{tab3}). At higher target velocities, the gains $K$ and $K_T$ are increased to enable the hexa-rotors to quickly reach the target, leading to a fast response from the hexa-rotors and oscillatory behavior about the desired hexa-rotor-target distance (see Case 3, Table \ref{tab2}). For the circular motion of the target, we tune the gains by increasing $K$ and decreasing $K_T$ such that the hexa-rotors can track the target while maintaining the desired separation (see Case 3, Table \ref{tab3}). For a smaller inter-hexa-rotor bound, the gain $K$ is significantly reduced to ensure that the hexa-rotors maintain the desired safe bounds (see Case 4, Tables \ref{tab2}, \ref{tab3}). {The control inputs eventually converge to the target velocity for the straight-line motion of the target. For circular motion of the target, the control inputs and UAV-target distance errors show a periodic nature, which is attributed to the associated target acceleration. The bounds on the inter-UAV distance tracking provide the necessary margin to accommodate for sensor noise and unmodeled dynamics in the control design and analysis while preserving the cyclic order of the UAVs in the formation. The cyclic order of the UAVs and target in the formation is preserved with BLF-control in both the cases of target motion.} The results of BLF-control for Case 1 and for the straight and circular motion of the target are presented in Figs. \ref{ROS_sim_circle_straight} and \ref{str-circ-ros-traj}. The ROS and numerical simulation videos are available at \url{https://youtu.be/-mXJm23ZQmE}.

\begin{figure}[t]
	\centering
	\includegraphics[height = 4.0cm]{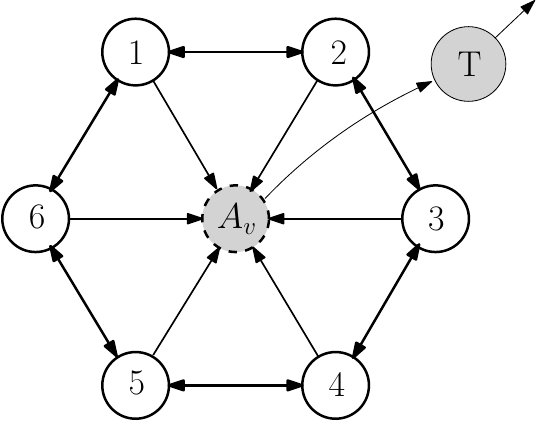}
	\caption{A system of $N = 6$ agents tracking a virtual agent with distance bounds, as the virtual agent tracks a target. Directional arrows indicate tracking of an agent; bi-directional arrows indicate mutual tracking of agents}
	\label{ma_trk_prb}
\end{figure}

\subsection{Application and Extension of the Problem}
The ROS implementation results of control \eqref{blf_control} applied to a two-hexa-rotor target tracking application can be extended to a problem with multiple hexa-rotors. Such a target tracking application would require a \textit{ring} topology for the agents to surround the target. Further, assuming a virtual agent ($\text{A}_{\text{v}}$) to be located at the centroid of this formation, the entire formation can be made to track a physical target (T) by making $\text{A}_{\text{v}}$ track T (see Fig. \ref{ma_trk_prb}). Additionally, by changing the function $U(\pmb{x},\pmb{x}_{\text{A}_\text{v}})$ \eqref{target_lyapunov} to a BLF, the \textit{ring} formation with the virtual agent will be preserved (assuming the virtual agent's velocity information is available to the surrounding agents).

{The BLF-control performs well for small disturbances and measurement noise. The control assumes that the inter-agent distances are within specified bounds for all $t \geq 0$. However, disturbances, measurement noise, and time delays may cause the bounds to be transgressed. As the system operates close to the distance bounds, a significantly high control magnitude (gradient) is required to keep the system operating within the bounds. Since the control on physical systems saturates, they may fail to keep the system within bounds, which get violated. A significant transgression accompanied by the saturation of control may lead to the failure of the control to stabilize the inter-agent distances to the specified distances. Thus, selecting the distance bounds (see \eqref{dist-bnd-sel}) based on system dynamics, disturbance and noise characteristics is vital to implementing BLFs on physical systems.}

\section{Conclusions and Future Work}\label{conc_sec}
Safe formation operation is an essential requirement of many proposed multi-UAV surveying and tracking applications. We presented a BLF-based control law to safely bound the inter-agent distances in a formation as applied to a collaborative target tracking application. The Lyapunov stability of the proposed control law was proved, and the formulation of a formation as a collection of bounded inter-agent distances was presented. This formulation can be easily extended to physical multi-UAV formations. {A detailed comparison of BLFs and QLFs was also presented. The advantages of using BLFs over QLFs in constraining the inter-agent distances and preserving the formation were illustrated using numerical simulations for the straight-line and circular motion of the target. The inclusion of distance and velocity measurement noise in these simulations validated the robustness of the BLF-based control in the formation control problem.} 

The extensive ROS environment simulations presented for a pair of hexa-rotors tracking targets executing constant velocity and circular motion provided valuable insights about the implementation of the algorithm in real-world applications, the shortcomings, and possible solutions to address them. The extension of the algorithm to applications with multiple hexa-rotors and obstacle avoidance was also discussed and is in scope for future work. A further analysis on the effects of control input saturation, quantifiable guarantees on steady-state errors, and undesired equilibria is also planned. Associated ROS environment simulations and eventual implementation of the algorithms on a team of multi-rotors will help in further evaluation of the advantages of the control proposed in this paper.

\section*{Declarations}

\textbf{Funding:} The authors did not receive support from any organization for the submitted work.\\
\textbf{Conflict of interest/Competing interests:} The authors have no relevant financial or non-financial interests to disclose.\\

\bibliographystyle{SageV}
\bibliography{sn-bibliography.bib}

\end{document}